\documentclass[runningheads]{llncs}
\usepackage{graphicx}
\usepackage{array}
\usepackage{multirow}
\usepackage{amssymb,enumitem,amsmath}
\usepackage{booktabs}
\usepackage{mathtools}
\usepackage{hyperref}
\usepackage{comment}
\usepackage{amsmath}
\usepackage{latexsym}
\usepackage{bbding}
\begin{document}
\title{Background Segmentation for Vehicle Re-Identification}
%
%
\author{Mingjie Wu\inst{1} \and
Yongfei Zhang\inst{1,2(}\Envelope\inst{)}  \and
Tianyu Zhang\inst{1} \and
Wenqi Zhang\inst{1}}
\authorrunning{F. Author et al.}
%
\institute{Beijing Key Laboratory of Digital Media, School of Computer Science and Engineering, Beihang University, Beijing 100191, China \\
\email{yfzhang@buaa.edu.cn} \and
State Key Laboratory of Virtual Reality Technology and Systems, Beihang University, Beijing 100191, China }
\maketitle              
\begin{abstract}
Vehicle re-identification (Re-ID) is very important in intelligent transportation and video surveillance.
Prior works focus on extracting discriminative features from visual appearance of vehicles or using visual-spatio-temporal information.
However, background interference in vehicle re-identification have not been explored.
In the actual large-scale spatio-temporal scenes, the same vehicle usually appears in different backgrounds while different vehicles might appear in the same background, which will seriously affect the re-identification performance. To the best of our knowledge, this paper is the first to consider the background interference problem in vehicle re-identification. 
We construct a vehicle segmentation dataset and develop a vehicle Re-ID framework with a background interference removal (BIR) mechanism to improve the vehicle Re-ID performance as well as robustness against complex background in large-scale spatio-temporal scenes. 
Extensive experiments demonstrate the effectiveness of our proposed framework, with an average 9\% gain on mAP over state-of-the-art vehicle Re-ID algorithms.
\keywords{Vehicle re-identification  \and Background segmentation \and Triplet loss.}
\end{abstract}
\section{Introduction}
Vehicle re-identification (Re-ID) targets to retrieve images of a query vehicle in different scenes~\cite{Liu2016Deep}.
In recent years, the number of cars has increased rapidly, and the application scenarios of intelligent transportation and video surveillance have pervasively expanded~\cite{liuu2016Deep}.
As one of the most important techniques therein, vehicle Re-ID has become a popular field, attracting widespread attention from both academia and industry.

\begin{figure}
\includegraphics[width=\textwidth]{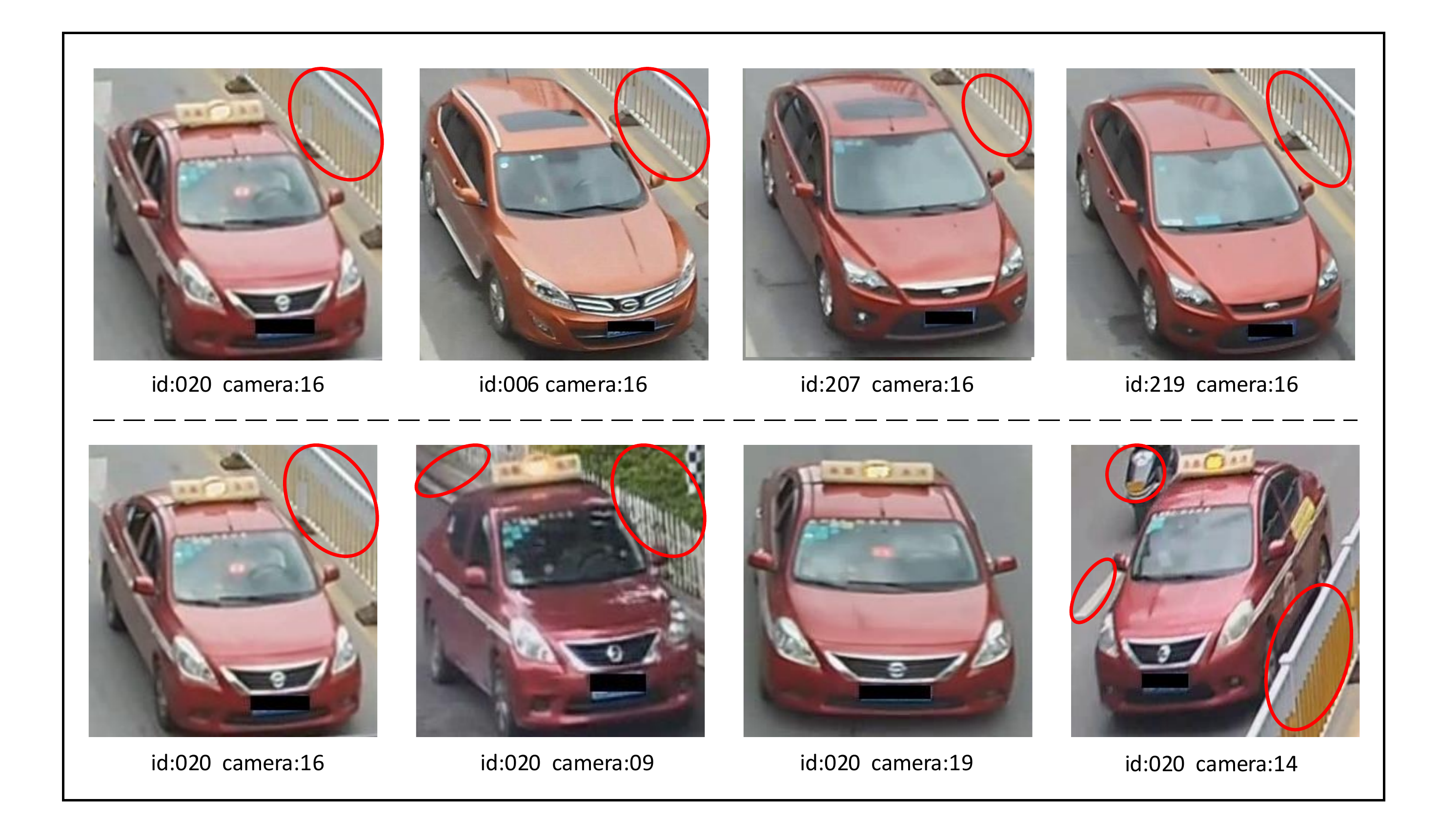}
\caption{The background interference information in datasets~\cite{Liu2016Large}. The first row of pictures shows the different vehicles shot under the same camera. The second row shows the pictures taken for the same car under different cameras.} \label{fig1}
\end{figure}

Prior works focus on visual appearance of vehicles~\cite{liu2018ram,Wang2017Orientation,Zhouy2018Viewpoint} only or introduce spatial-temporal clues for more accurate vehicle Re-ID~\cite{Chih2018Vehicle,Shen2017Learning}.
With the nice exploration of visual appearance of vehicles and/or spatial-temporal constraints, as well as the carefully designed deep learning architectures, the performance of vehicle Re-ID has been largely improved, yet there are still unexplored issues.
One of them is the background interference problem.
In the actual large-scale scenes, the same vehicle often appears in different backgrounds and different vehicles may appear in the same background, which will seriously affect the Re-ID performance, due to background interference.
As shown in the first row of Fig.~\ref{fig1}, different vehicles share the same background, making it harder to distinguish.
The second row shows the same car in different cameras.
The various background may lead Re-ID systems to extract background-related features and result in unsatisfying Re-ID performance.

In this paper, we propose to use vehicle background segmentation to reduce the harmful impact of interference information and improve the vehicle Re-ID performance as well as robustness against complex background in large-scale spatio-temporal scenes.
The extensive experiments demonstrate that the proposed framework can accurately extract the vehicle characteristics and outperform several state-of-the-art methods. 
Our contributions are summarized as follows:
\begin{itemize}
\item
We construct a vehicle background segmentation dataset based on the Citysc- apes~\cite{cordts2016cityscapes} dataset, and then use the new dataset to train the vehicle background segmentation model.
\item
We propose a novel post processing method for more accurate background segmentation. The post processing procedure ensures the segmented images reserve the vehicle area as much as possible, thus the segmentation suffers less information loss.
\item
We develop a new background interference removal (BIR) pipeline and evaluate our methods on two large-scale vehicle Re-ID datasets~\cite{liuu2016Deep,Liu2016Large}. Experiment results demonstrate that by reducing background interference, the Re-ID performance can be boosted significantly.
\end{itemize}

\section{Related Works}
Vehicle Re-ID is an essential task in the field of intelligent transportation and video surveillance. In this section, we first review prior works on vehicle Re-ID. Since we apply background segmentation, we also study previous detection and segmentation methods. Finally, we review the loss functions for embedding.
\subsection{Vehicle Re-identification}
Inspired by person Re-ID, vehicle Re-ID task has attracted much attention in the past few years. RAM~\cite{liu2018ram} extracts global features and local regions feature. As each local region conveys more distinctive visual cues, RAM encourages the deep model to learn discriminative features. The confrontation training mechanism and the auxiliary vehicle attribute classifier are combined to achieve efficient feature generation. Zhou~\emph{et al.}~\cite{Zhouy2018Viewpoint} proposed a viewpoint-aware attentive multi-view inference (VAMI) model. Given the attentional features of single-view input, this model designs a conditional multi-view generation network to infer the global features of different viewpoint information containing input vehicles. Liu~\emph{et al.}~\cite{Liu2016Large} released the VeRi-776 dataset, in which there exist more view variants. And they proposed a new mathod named FDA-net\cite{lou2019veri}. They use visual features, license plates and spatio-temporal information to explore Re-ID tasks. 
In addition, Shen~\emph{et al.}~\cite{Shen2017Learning} and Wang~\cite{Wang2017Orientation} respectively proposed visual-space-time path suggestion and spatio-temporal regularization, focusing on the development of vehicle spatio-temporal information to solve vehicle Re-ID problem.

\subsection{Target Detection and Background Segmentation}
Conventional target detection systems utilize classifiers to perform detection.
For example, Dean~\emph{et al.}~\cite{dean2013fast} adopts the method of deformable parts models (DPM), and the recent R-CNN~\cite{girshick2014rich} method uses region proposal methods. The flow of such methods is complicated, and there are problems of slow speed and difficult training. The YOLOv3~\cite{Redmon2018YOLOv3} detection system runs a single convolutional network on the image, and the threshold of the obtained detection is dealt by the confidence of the model. The detection is quite quick and the process is simple. YOLOv3 can utilize the full map information in the training and prediction process. Compared to Fast R-CNN~\cite{girshick2015fast}, the YOLOv3 background prediction error rate is half lower.

With the application of convolutional neural networks (CNN)~\cite{sharif2014cnn} to image segmentation, there has been increasing interest in background pixel annotation using CNN with dense output, such as F-CNN~\cite{zhao2016f}, deconvolution neural network~\cite{noh2015learning}, encoder decoder SegNet~\cite{badrinarayanan2017segnet}, and so on. DilatedResNet~\cite{Yu2015Multi} and PPM~\cite{Zhao2016Pyramid} have made great progress in the field of background segmentation. Compared with other CNN-based methods, this method has higher applicability and stable performance.

\subsection{Loss Functions for Embedding}
Hermans~\emph{et al.}~\cite{Hermans2017In} demonstrated that the use of triplet-based losses to perform end-to-end depth metric learning is effective for person Re-ID tasks. Kanac~\cite{kanaci2017vehicle} proposed a similar method for vehicle embedding based on the training model of vehicle model classification. Cross entropy loss ensures separability of features, but these features may not be distinguishable enough to separate identities that are unseen in the training data.
Some recent works~\cite{rippel2015metric,shen2016relay,wojke2018deep} combined the classification loss through metric learning. Kumar~\emph{et al.}~\cite{kumar2019vehicle} conducted an in-depth analysis of the vehicle triplet embeddings, extensively evaluated the loss function, and proved that the use of triplet embeddings is effective.

\section{Approach}

Since there is no background segmentation dataset specially designed for vehicle segmentation, we first explain how we use the target detection algorithm to construct the dataset and then we introduce the BIR module and the overall framework.

\subsection{Constructing Vehicle Background Segmentation Dataset}

Given the powerful performance and good stability of Yolov3, we use Yolov3 as a target detection tool. 
The Cityscapes dataset is powered by Mercedes-Benz and provides image segmentation annotations. Cityscapes contains 50 scenes of different views, different backgrounds and different seasons.
Because they are high definition images and most of them are street view pictures, this dataset image is suitable for further cropping as a vehicle background segmentation dataset.

\begin{figure}
\includegraphics[width=\textwidth]{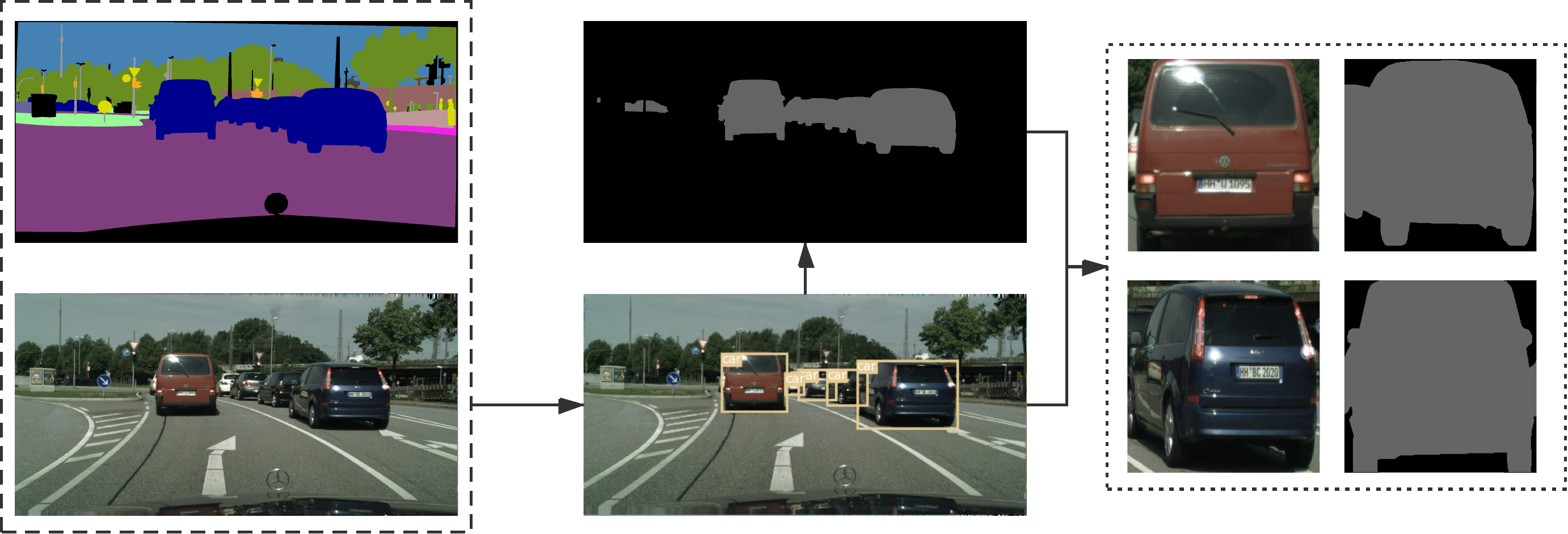}
\caption{Generating CS-vehicle. On the left is the Cityscapes dataset. We use yolov3 on this dataset to detect the vehicles (middle), remove the detected vehicles with less than $256\times256$, and then generate the images we need (right).} \label{fig2}
\end{figure}

When adopting Yolov3 for vehicle detection, we ignore the detected vehicle images smaller than $256\times256$ because the small images are unclear and affect the segmentation effect.
And we generated the small dataset named Cityscapes-vehicle (CS-vehicle for short) which contains cropped 2,686 vehicle images and only vehicle segmentation annotations as well as two labels (the vehicle and the background), as shown in Fig. \ref{fig2}

However, CS-vehicle has rather less images, so we construct a large vehicle background segmentation dataset named Vehicle-Segmentation to train the vehicle background segmentation model.
The Vehicle-Segmentation contains a total of 27,896 images: 2,686 vehicle images from CS-Vehicle, 5,000 street view images from Cityscapes and 20,210 all kinds of scenes images from ADE20K~\cite{Zhou2017Scene}.
Using Cityscapes and ADE20K are aimed to extract more semantic features, while using CS-Vehicle can make our model pay more attention on vehicle segmentation.
All of the annotated images in Vehicle-Segmentation have a resolution greater than 256*256, which is very suitable for background segmentation tasks in vehicle Re-ID.

\subsection{Vehicle Re-ID with Background Inteference Removal}
\begin{figure}
\includegraphics[width=\textwidth]{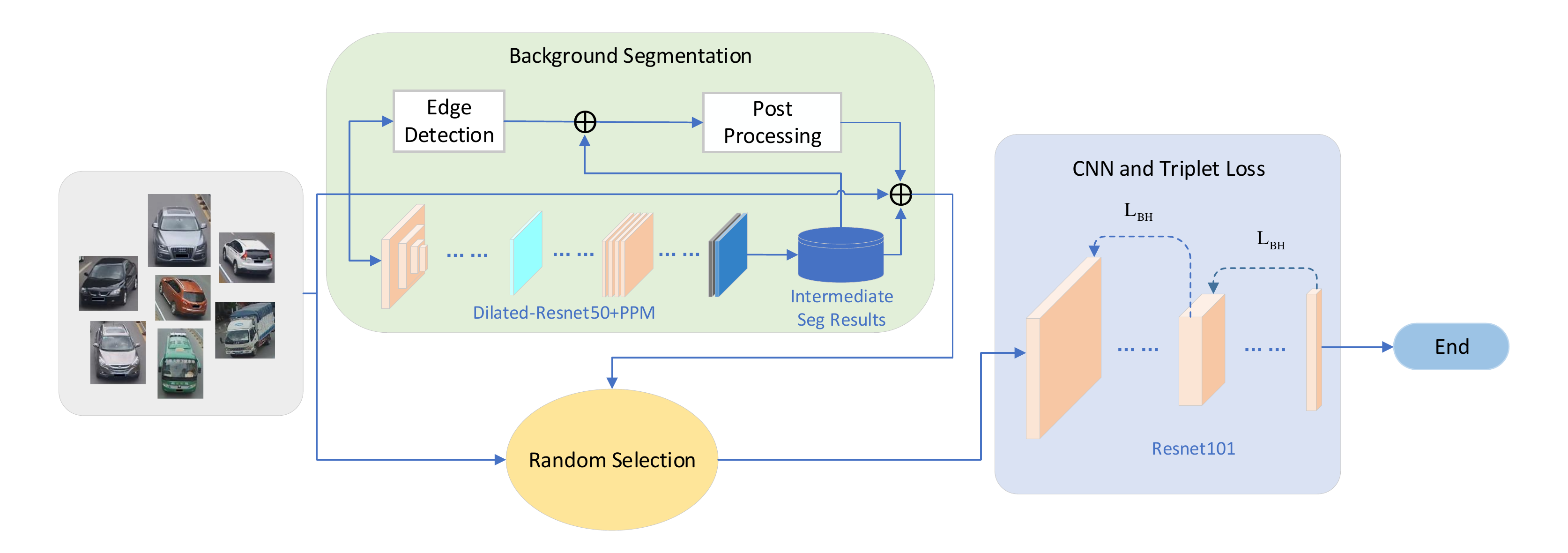}
\caption{Structure of the Vehicle Re-ID with background interference removal (BIR), which is composed of three components. Background Segmentation, Random Selection, CNN and Loss Function.} \label{fig3}
\end{figure}

In this section, we introduce a vehicle Re-ID framework with background interference removal (BIR) as illustrated in Fig.~\ref{fig3}. 
Our framework consists of three components, a background segmentation module to remove background interference, a random selection module to enhance robustness and a CNN-Loss module to achieve Re-Id tasks. We will explain the details of the three main components in this section.

\subsubsection{Background Segmentation}
Based on the original semantic segmentation method, we find that the background segmentation results are unsatisfying, which affects the vehicle Re-ID accuracy seriously when directly used to train the Re-ID network.
This is due to the presence of uncomplete segmented vehicle bodies with holes and interference factors such as labels, license plates, and faces.
Therefore, we propose a post processing method for more accurate background segmentation. 
We use the DilatedResNet-50 network as the encoder and the PPM network as the decoder for the initial scene segmentation to generate the intermediate segmentation results.
Then we combine the edge detection results with intermediate segmentation results, and we further process them with our post processing method.
First, the flood water filling algorithm are used to fill the holes; second, we detect the number of connected regions in the binary image and mark each connected domain, leaving only the largest connected domain; finally, we discard those segmented images in which the car contains a pixel area smaller than $0.60$ of the overall picture area.
\subsubsection{Random Selection}
Considering the fact that vehicle background segmentation effect is not perfect, we design a random selection module in our framework. 
All images has a probability of $k$ ($k\in[0,1]$) to be selected to use their images with background segmentation. These segmentation images, together with other $1-k$ ratio original images, are put into CNN to train the Re-ID module. 

By using this random selection module, we reduce the risk of imperfect segmentation results and increase the diversity of vehicle images. 
This further encourages the model to extract more discriminative features from vehicle body areas and improve the generalization ability.

\subsubsection{Loss Function}

The training images are first fed into a CNN backbone to produce feature vectors.
In view of the excellent characteristics of batch hard triplet loss in the field of person Re-ID, we use it for vehicle re-identification.

To reduce the number of triples, the authors~\cite{Hermans2017In} construct a data batch by randomly sampling $P$ identities and then randomly sampling $K$ images for each identity, thus resulting in a batch size of $PK$ images.
For each sample in the batch, the most difficult positive sample and the most difficult negative sample in this batch are selected when forming the triplet for calculating the loss, as shown in Eq. 1. 
\begin{equation}
\begin{split}
    \mathcal{L}_{BH}(\theta)=\sum_{i=1}^P\sum_{a=1}^K [\max_{\substack{p=1...K}} D(f_{\theta}(x^a_i),f_{\theta}(x^p_i))- \\
   \min_{\substack{n=1...K \\ j=1...P\\j \neq i} }  D(f_{\theta}(x^a_i),f_{\theta}(x^n_j))+m]_{+} \\
\end{split}
\end{equation}
where $\theta$ is the parameter that model learns in deep neural network, $f_\theta$ is the feature vector of image, $x^j_i$ corresponds to the $j$-th image of the $i$-th vehicle in the batch. $a$ denotes the \textit{anchor}, $p$ denotes the \textit{positive}, $n$ denotes the \textit{negative} and $D$ denotes the distance between two feature vectors.

\section{Experiment}
\subsection{Datasets}
\subsubsection{VeRi-776}
VeRi-776 dataset is collected with 20 cameras in real-world traffic surveillance environment. A total of 776 vehicles are annotated. 200 vehicles are used for testing. The remaining 576 vehicles are for training. There are 11,579 images in the test set, and 37,778 images in the training set.    

\subsubsection{VehicleID}
VehicleID dataset is captured by multiple surveillance cameras and there are 221,763 images of 26,267 vehicles in total (8.44 images/vehicle in average). This dataset is split into two parts for model training and testing. The first part contains 110,178 images of 13,134 vehicles and 47,558 images have been labeled with vehicle model information. The second part contains 111,585 images of 13,133 vehicles and 42,638 images have been labeled with vehicle model information. This dataset extracts three subsets, \emph{i.e.} small, medium and large, ordered by their size from the original testing data for vehicle retrieval and vehicle Re-ID tasks.

\subsection{Implementation Details}
We use the ResNet-101 architecture and the pre-trained weights provided by He \emph{et al.}~\cite{he2016deep}. 
In a mini-batch, we randomly select 18 vehicles and for each vehicle, we randomly select 4 images, thus the batch size is 72.
We run 300 epochs with a base learning rate $2\times10^{-4}$.
The input image size is set to $256\times256$.
We use two GTX 2080ti GPUs for training.

\begin{figure}
\includegraphics[width=\textwidth]{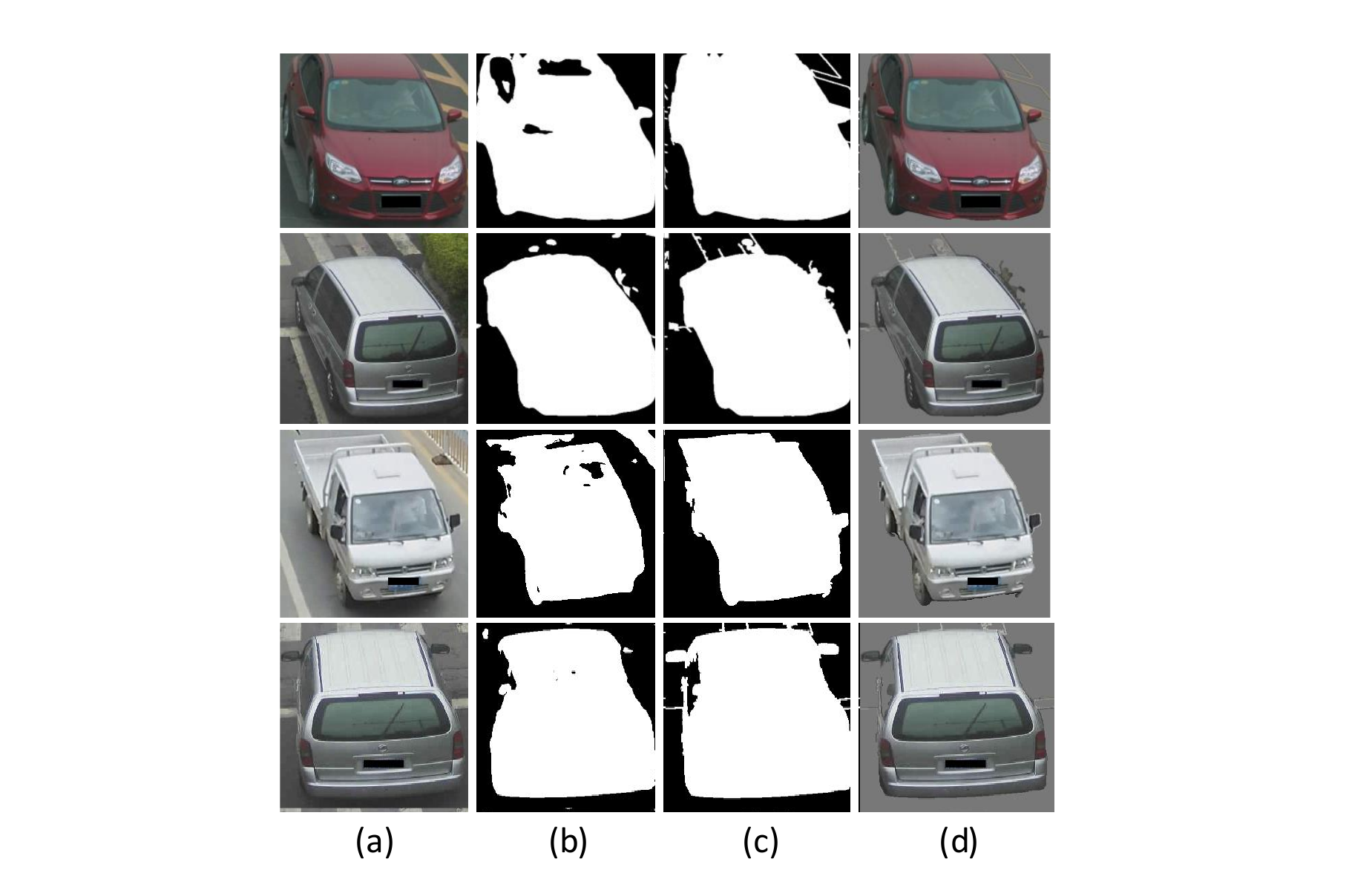}
\caption{The vehicle segmentation results. (a) original image, (b) intermediate segmentation results, (c) our background removal results, (d) final results.} \label{fig4}
\end{figure}
\subsection{Performance Evaluation}
In this section, we show the effect of background segmentation, experiment results of BIR and comparisons with state-of-the-art methods. 
The baseline model involves no segmentation.
The evaluation is conducted in four steps:
\begin{itemize}
\item
\textit{Seg} first trains the model only using the dataset processed by the Dilated-Resnet50+PPM module.
\item
\textit{Seg+Post} further adds the post processing to the background removal module then uses the new dataset to train vehicle Re-ID model.
\item
\textit{TrainS+TestN} uses the training dataset that processed by \textit{Seg+Post} as new training dataset and uses original image without any processing as test dataset.
\item
\textit{Random-k} randomly uses $k$ percent images processed by \textit{Seg+Post} and others original images for both training and test to evaluate the vehicle Re-ID model.
\end{itemize}

\subsubsection{Effect of Background Segmentation}

The vehicle background removal results are shown in Fig. \ref{fig4}. It is obvious that our background removal results are better than preliminary results.
We show the improvements gained by improving background removal methods and the experimental results on VeRi-776 are summarized in Tab. \ref{tab1}. We found that the results of \textit{Seg} have not been improved, but have been reduced. We believe that the segmentation effect is not good. It can be seen that the effect of \textit{Seg+Post} is significantly improved. Therefore, background removal is effective for the vehicle to identify. \textit{TrainS+TestN} shows the effectiveness of  background interference removal.\par
\begin{table*}[]
\centering
\caption{Results(\%) of ablation experiment on VeRi-776}
\label{tab1}

\begin{tabular}{p{5cm}<{\centering} p{1.5cm}<{\centering} p{1.5cm}<{\centering} p{1.5cm}<{\centering} p{1.5cm}<{\centering}}
\toprule
Method &  mAP & top1 & top5 & top10\\
\midrule
Seg & 57.53 & 81.70 & 82.01 & 94.76 \\
Seg+Post & 64.79 & 86.89 & 95.47 & 97.85 \\
TrainS+TestN& 66.08 & 86.89 & 94.87 & 97.14 \\
\midrule
Baseline  & 65.78 & 86.29 & 94.76 & 96.96 \\
\bottomrule
\end{tabular}
\end{table*}

\subsubsection{Experiment Result of BIR}
In this section, we experimentally verify the impact of the parameter $k$ on the performance and determine the best one accordingly. 
As shown in Tab. \ref{tab2}, when the $k$ is equal to 0.2, the effect is best.  
It exhibits a substantial improvement of 4.96\% in mAP over the baseline method that uses the original images rather than the segmented images to train and test.
All of our results have better effect than baseline. 
But the effect is not significant as $k$ increases. The reason may be that the segmentation performance is not perfect. 

\begin{table*}[]


\centering
\caption{Results(\%) of BIR experiment on VeRi-776}
\label{tab2}

\begin{tabular}{p{5cm}<{\centering} p{1.5cm}<{\centering} p{1.5cm}<{\centering} p{1.5cm}<{\centering} p{1.5cm}<{\centering}}
\toprule
k &  mAP & top1 & top5 & top10\\

\midrule
0.1 & 70.12 & 90.28 & \textbf{97.08} & \textbf{98.81} \\
0.2 & \textbf{70.74} & \textbf{90.46} & 96.96 & 98.69 \\
0.3 & 69.61 & 89.03 & 95.89 & 98.15 \\
0.4 & 70.10 & 89.98 & 96.48 & 98.15 \\
0.5 & 69.65 & 89.27 & 96.42 & 98.51 \\
0.6 & 68.59 & 88.67 & 96.00 & 98.15 \\
0.7 & 70.34 & 89.45 & 96.06 & 98.39 \\
0.8 & 67.69 & 88.73 & 95.59 & 97.79 \\
0.9 & 67.22 & 87.72 & 95.89 & 97.79 \\
\midrule
Baseline  & 65.78 & 86.29 & 94.76 & 96.96 \\
\bottomrule
\end{tabular}
\end{table*}

\subsubsection{Comparisons with State-of-the-art Methods}

As shown in Tab. \ref{tab3}, our vehicle Re-ID framework with BIR outperforms state-of-the-art methods and compared baselines, which demonstrates the effectiveness of our overall framework and individual components.
Compared with \textit{FDA-net}~\cite{lou2019veri}, BIR has a gain of 15\% in terms of mAP and 6\% in terms of top-1 accuracy. Such a performance increase shows that the batch hard triplet loss does provide vital priors for robustly estimating the vehicle similarities. 
Compared with \textit{MOV1+BH}~\cite{kumar2019vehicle} only, BIR also has a 5.6\% increase in terms of mAP. This is because that the our method eliminates background interference information. It strongly proves that background segmentation is of great significance in vehicle Re-ID. 

Although BIR only uses visual information, our final approach also has a 17\% and 12\% increase in terms of mAP, as compared with Siamese-CNN+path-LSTM~\cite{Shen2017Learning} and AFL+CNN~\cite{Chih2018Vehicle}, which also use spatio-temporal information. It proves that making full use of image information can improve accuracy.
Compared with the \textit{VAMI}~\cite{Zhouy2018Viewpoint} and \textit{RAM}~\cite{liu2018ram}, BIR is not complicated and we don't have to pick a dataset image and our method has a 9\% mAP gain \par

It should be noted that, BIR does not use local information features, nor does it add spatio-temporal information, and only considers the global feature, and the performance is greatly improved. This shows that the background of the picture has a great interference to vehicle Re-ID and removing the background is of great significance for vehicle Re-ID.

\begin{table*}[!ht]


\centering

\caption{ Comparisons (\%)  with State-of-the-art Re-ID methods on VeRi-776 and VehicleID}

\label{tab3}

\begin{tabular}{ccccccccc}

\toprule
\multirow{1}{*}{Dataset} & \multicolumn{2}{c}{VeRi} & \multicolumn{6}{c}{VehicleID}\\
\midrule
\multirow{2}{*}{Method} & \multicolumn{2}{c}{all} & \multicolumn{2}{c}{small} & \multicolumn{2}{c}{medium} & \multicolumn{2}{c}{large} \\

\cmidrule(r){2-3} \cmidrule(r){4-5} \cmidrule(r){6-7} \cmidrule(r){8-9}

&  mAP  &  top1      

&  mAP  &  top1      

&  mAP  &  top1      

&  mAP  &  top1        \\

\midrule

MOV1+BH\cite{kumar2019vehicle}		& 65.10  & 87.30 
                                    & 83.34  & \textbf{77.90}    
									& 78.72  & 72.14 
									& \textbf{75.02}  & 67.56     \\
									
RAM~\cite{liu2018ram}		        & 61.50 & 88.60 
                                    &  ---  & 75.20  
									&  ---  & 72.30  
									&  ---  & \textbf{67.70}    \\ 									
									 			
FDA-Net\cite{lou2019veri}		    & 55.49 & 84.27 
                                    &  ---  &  ---   
									& 65.33 & 59.84  
									& 61.84 & 55.53    \\

Siamese-CNN+path-LSTM\cite{shen2016relay}	& 58.27 & 83.94 	
                                            &  ---  &  ---  
									        &  ---  &  ---   
									        &  ---  &  ---    \\ 
AFL+CNN\cite{Chih2018Vehicle}		&53.35 & 82.06 
                                    &  ---  &  ---     
									&  ---  &  ---    
									&  ---  &  ---     \\ 

VAMI~\cite{Zhouy2018Viewpoint}		&61.32 & 85.92 
                                    &  ---  & 63.12   
									&  ---  & 52.87 
									&  ---  & 47.34    \\ 

\midrule
Our Method	&\textbf{70.74} & \textbf{90.46}  	
& \textbf{85.46} & 77.17 
& \textbf{84.41} & \textbf{75.81} 
& 74.13 & 63.71    \\

\bottomrule

\end{tabular}

\end{table*}

\subsection{Discussion}
Our proposed BIR is suitable for pictures taken under different cameras across large spatial-temporal span.
During the experiment, we find that background removal results for white vehicles are better than that for black vehicles, because the road is mostly dark gray, thus black vehicles are more difficult to segment.
We will consider this effect in the background removal, and optimize to generate a more widely used algorithm in the future work.

\section{Conclusion}
In this paper, we proposed a new vehicle Re-ID framework with background interference removal to improve the generalization ability over large spatio-temporal span. Due to the absence of vehicle background segmentation datasets, we construct a vehicle background segmentation dataset based on Cityscapes, and then use this dataset for the training of vehicle background segmentation models. In the later stage of segmentation, we use the traditional image operation to post-process the segmentation results, and generate a vehicle image with better background removal. We use proposed BIR for the Re-ID task, and the vehicle Re-ID accuracy is significantly improved, outperforming state-of-the-art vehicle Re-ID algorithms. It should also be noted that different from existing methods, which either introduced local visual information or spatio-temporal information, the proposed scheme uses only the global visual information of vehicles, which can be easily integrated  with these algorithms to further enhance the vehicle Re-ID accuracy. 

\subsubsection{Acknowledgments.}
This work was partially supported by the National Natural Science Foundation of China (No. 61772054), and the NSFC Key Project (No. 61632001) and the Fundamental Research Funds for the Central Universities. 

\bibliographystyle{splncs04}
\bibliography{bibfile}

\end{document}